\documentclass[runningheads]{llncs}
\usepackage{graphicx}
\usepackage{amsmath}
\usepackage{multirow}
\usepackage{colortbl}
\usepackage{xcolor}
\usepackage{color}
\usepackage{booktabs}
\newcommand{\RNum}[1]{\uppercase\expandafter{\romannumeral #1\relax}}
\begin{document}
\title{WaveFuse: A Unified Unsupervised Framework for Image Fusion with Discrete Wavelet Transform}
\titlerunning{WaveFuse: A Unified Image Fusion Framework}
%
\author{Shaolei Liu \inst{1,2} \and
Manning Wang\inst{1,2} \thanks{Corresponding authors.} \and
Zhijian Song\inst{1,2}$^\star$ }
\authorrunning{S. Liu, M. Wang, Z. Song et al.}
%
\institute{Digital Medical Research Center, School of Basic Medical Science, Fudan University, China \and
Shanghai Key Laboratory of Medical Image Computing and Computer Assited Intervention\\
\email{{slliu, mnwang, zjsong}@fudan.edu.cn}}

\maketitle              
\begin{abstract}
    We propose an unsupervised image fusion architecture for multiple application scenarios based on the combination of multi-scale discrete wavelet 
    transform through regional energy  and deep learning. To our best knowledge, this is the first time that
    a conventional frequency method has been combined with deep learning for feature maps fusion. The useful information 
    of feature maps can be utilized adequately through multi-scale discrete wavelet transform in our proposed method.
    Compared with other state-of-the-art fusion methods, the proposed algorithm exhibits better fusion 
    performance in both subjective and objective evaluation. Moreover, it's worth mentioning that 
    comparable fusion performance trained in COCO dataset can be obtained by training with a much smaller
    dataset with only hundreds of images chosen randomly from COCO. Hence, the training time is shortened 
    substantially, leading to the improvement of the model's performance both in practicality and training 
    efficiency.
\end{abstract}

\section{Introduction}

Image  fusion is  the technique of integrating complementary information from multiple 
images obtained by different sensors of the same scene, so as to improve the  richness of 
the information contained in one image \cite{goshtasby2007imagesota}. Image fusion can compensate for
the limitation of single imaging sensors, and this technique has developed rapidly in recent years 
because of  the wide availability of different kinds of imaging devices \cite{goshtasby2007imagesota}. 
For example, in medical imaging applications, images of different modalities can be fused  to achieve more reliable and 
precise medical diagnosis \cite{song2019msdnetMEDICAL}. In military surveillance applications, image fusion 
integrates information from different electromagnetic spectrums (such as visible  and infrared bands) 
to achieve night vision \cite{li2017pixelsota}.


  
The extraction of feature maps and the selection of fusion rules are the two key
factors determining the quality of the fused image \cite{liu2017multifocus}, and most studies focus on proposing new methods 
based on  these two factors.
  Before the overwhelming application of deep learning in image processing, many conventional approaches were
   used in feature extraction for image fusion, which can be divided into two categories: transform domain 
   algorithms and spatial domain algorithms \cite{li2017pixelsota}. 
  In transform domain algorithms, the source images are transformed to a specific transform domain, such
  as  the frequency domain, where the feature maps are represented by the decomposition coefficients of 
  the specific transform domain. In feature maps fusion, max-rule and averaging are commonly used for high and low frequency 
  bands, respectively, and then the fused image is reconstructed by the inverse transform from the fused 
  feature maps \cite{li1995DWT,NSCT}. Unlike transform domain algorithms, spatial domain algorithms employ the original pixel of 
  source images as feature maps and directly calculate the weighted average of the source images to obtain 
  the final fused image without dedicated feature maps extraction, where the weights are selected according
   to image blocks \cite{imageblocks}  or gradient information \cite{gradientinfo}. Consequently,
   the conventional approaches can be regarded as designing some hand-crafted filters 
   to process the source images, and it is difficult for them to adapt to images of different 
   scenes or parts with different visual cues in one image.

   Nowadays, deep learning has been the state-of-the-art solution in most tasks in the fields of image processing and computer vision.
  Recently, deep learning has also been used in image fusion and achieved higher quality than conventional methods. For example,  
  CNN can  be  used  to  automatically  extract  useful features and can learn the direct mapping from source images to  feature maps. 
  In recent image fusion research based on deep learning \cite{liu2017multifocus,prabhakar2017Deepfuse,li2018densefuse,song2019msdnetMEDICAL,du2017image,liu2018dppixel,li2019resnet}, 
  fusion using learned features through CNN achieved higher quality than conventional fusion approaches.
   According to the different fusion framework utilized, 
  deep learning based methods can be divided  into the following three categories: CNN based methods \cite{liu2017multifocus,liu2017medical,li2019resnet,11}, 
  encoder-decoder based methods \cite{prabhakar2017Deepfuse,li2018densefuse,song2019msdnetMEDICAL,12} and 
  generative adversarial network (GAN) based methods \cite{ma2019fusiongan}. CNN based methods 
  merely apply several convolutional layers to obtain the weight map for source images. Encoder-decoder based 
  methods introduce encoder-decoder architecture to extract deep features, and the deep features are fused by 
  weighted average or concatenation. Furthermore, GAN based methods leverage conditional GAN to generate the 
  fused image, where the concatenated source images are directly input to the generator. In these studies, the 
  feature maps obtained through deep learning are usually simply fused by weighted averaging, and we will show 
  that this is not optimal.
  More importantly,
  the neural networks used in these studies \cite{liu2017multifocus,prabhakar2017Deepfuse,li2018densefuse,song2019msdnetMEDICAL,li2019resnet}
  usually need to be trained on large image dataset, which is 
  time consuming.

  In this paper, we propose an image fusion algorithm by combining the deep learning based 
  approaches with conventional transform domain based approaches. Concretely, we first train an encoder-decoder
  network and extract feature maps from the source images by  the encoder. Inspired by the multi-scale transform \cite{li2017pixelsota}, 
  discrete wavelet transform (DWT) is  utilized to transform the feature maps 
  into the wavelet domain, and adaptive fusion rules are used at low and high frequencies, thus  making the beat use of  the 
  information of feature maps. Finally, inverse wavelet 
  transform is used to reconstruct the fused feature map, which is decoded by the decoder to obtain the final fused 
  image. Experiments show that with the additional processing of the feature maps by DWT, the quality of the fused 
  image  is remarkably improved. To the best of our knowledge, this is the first time to adopt conventional transform 
  domain approaches to fuse the feature maps obtained from deep learning approaches.

  The main contributions are summarized as follows:

(1) A generalized and effective unsupervised  image fusion framework is proposed based on the combination of 
multi-scale discrete wavelet transform  and deep learning.

(2) With multi-scale decomposition in DWT, the useful information of feature maps can be fully utilized. 
Moreover, a region-based fusion rule is adopted to capture more detail information. Extensive
experiments demonstrate the superiority of our network  over the state-of-the-art fusion methods.

(3) Our network can be trained in a smaller dataset with low computational cost to achieve comparable 
fusion performance compared with existing deep learning based methods trained on full  COCO dataset.
 Our  experiments show that the quality of the 
fused images and the training efficiency are improved sharply.

\section{Proposed Method}

\subsection{Network Architecture}
\begin{figure*}[t]
  \centering
  \includegraphics[scale=0.36]{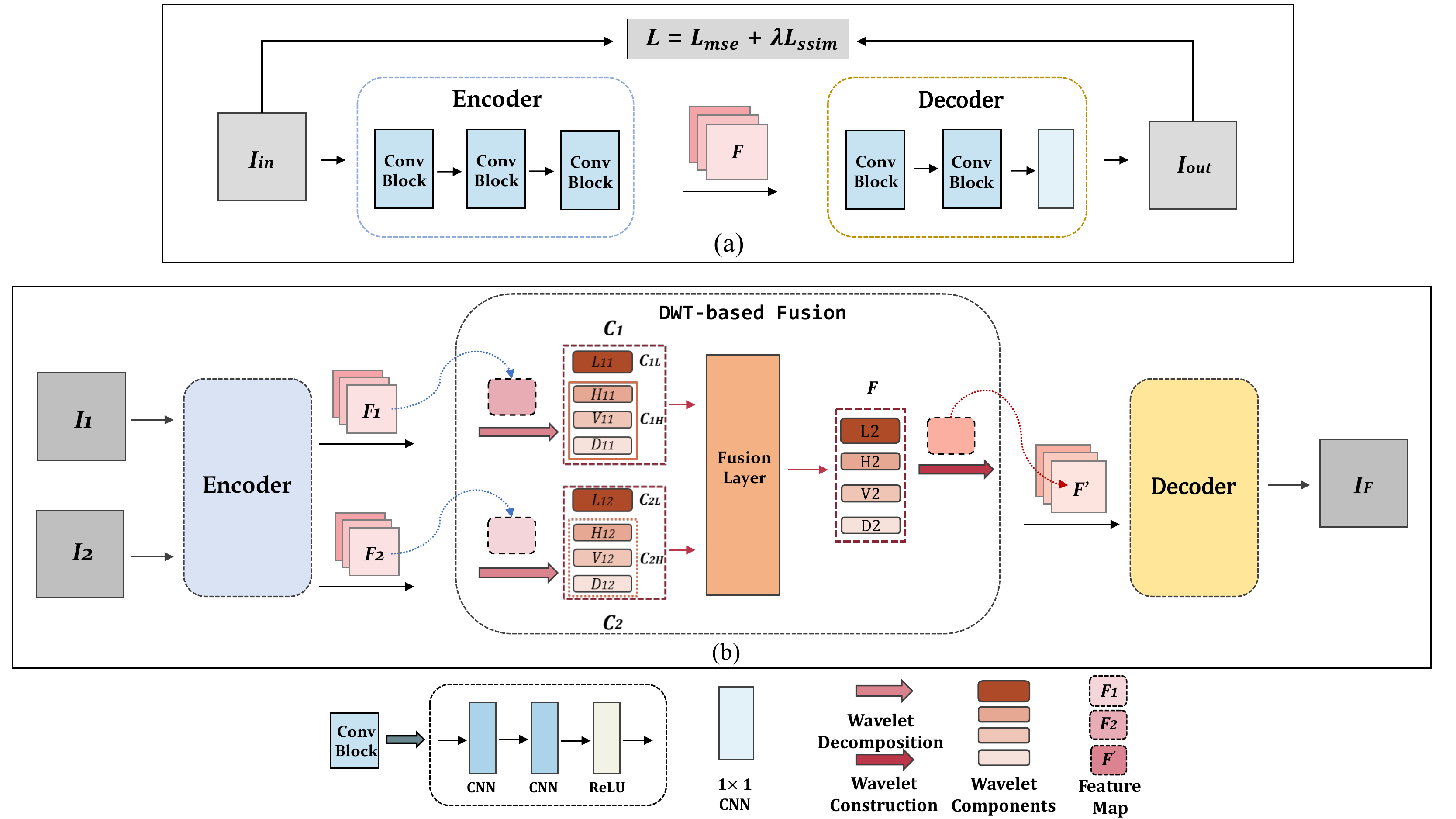}
  \caption{(a) The framework of the training process. (b)Architecture of the proposed WaveFuse image fusion network. The feature maps learned by the 
  encoder from the input images are processed by multi-scale discrete wavelet transform, and finally the fused 
  feature maps are utilized to the fused image reconstruction by the decoder.}
  \label{fig:net}
\end{figure*}

WaveFuse is a typical encoder-decoder architecture, consisting of three components: an encoder, a DWT-based
fusion part and a decoder.  As shown in Fig. \ref{fig:net} (b), the inputs of the network are spatially aligned source images $I_k$, 
where $k$ = 1,2 is used to index the images.  Feature maps $F_k$ are obtained by extracting features from 
the input source images $I_k$ through the encoder. The feature maps $F_k$ are first transformed into the
 wavelet domain, and an adaptive fusion rule is used to obtain the fused feature maps
 $F^{'}$. Finally, the fused feature maps are input into the decoder to obtain the final fused image  $I_F$. 
The encoder is composed of three ConvBlocks, where two CNNs and a Relu layer are included.  The 
kernel size of CNNs are all 3 $\times $ 3. After encoding, 48D feature maps are obtained for fusion.
In the DWT-based fusion part, to take 1 layer wavelet decomposition and one dimension of the feature maps $F_k$ for example,  the feature maps  are decomposed  to   different wavelet components $C_k$, 
including one  low-frequency component $C_{kL}$, namely $L_{1k}$ and three high-frequency 
components $C_{kH}$: horizontal component $H_{1k}$, vertical component  $V_{1k}$ and diagonal component $D_{1k}$, respectively.   
Different fusion rules are employed for different components to obtain the fused wavelet components $F$, where the low-frequency component
$L_2$  is obtained from the fusion of $L_{11}$ and $L_{12}$, and the high-frequency components $H_2$, $V_2$ and $D_2$ are obtained 
from the fusion of $H_{1k}$, $V_{1k}$ and $D_{1k}$, respectively. Finally, the fused low-frequency component and high-frequency 
components are integrated by wavelet reconstruction to obtain the final fused feature map $F^{'}$. 
The decoder is mainly composed of two ConvBlocks and one 1x1 CNN, where the 
fused image is finally reconstructed.

\subsection{Loss Function}
The loss function $L$ used to train the encoder and the decoder in WaveFuse  is a weighted combination of 
pixel loss $ L_p$ and structural similarity loss $L_{ssim}$ with a weight $\lambda $ , where  $\lambda $ is assigned as 1000 according to \cite{li2018densefuse}.
The loss function $L$, pixel loss $ L_p$ and structural similarity loss $L_{ssim}$  are  defined as follows:

\begin{small}
    \begin{align}
      L&=L_p+\lambda L_{ssim}, \label{loss}\\
      Lp&=|| I_{out}-I_{in}||_2, \\  
      L_{ssim}& = 1-SSIM(I_{out},I_{in}),
    \end{align}
  \end{small}

\noindent where $I_{in} $ and $I_{out}$ represent the input 
image to the encoder and the  output image of the decoder, respectively. The structural similarity  (SSIM) 
is a widely used perceptual image quality metric, which combines the three components of luminance, structure and 
contrast to comprehensively measure image quality \cite{wang2004ssim}.

\subsection{Training}

We trained our network shown in Fig.1 (a) using COCO \cite{coco} containing 70,000 images, and all of them were resized 
to 256 $\times$ 256 and transformed to gray images. The batch size and epochs were set as 64 and 50,
respectively. Learning rate was $1\times10^{-4}$. The proposed method was implemented on Pytorch 1.1.0 with Adam as the 
optimizer and a NVIDIA GTX 2080 Ti GPU for training. 
In our practical training process, we found that using comparatively small dataset, containing 300-700 images chosen 
randomly from COCO,  still achieved a comparable fusion quality. 
The  parameters for small dataset are as follows: learning rate was set as $1\times10^{-4}$, and the batch size 
and epochs were 16 and 100, respectively.

\subsection{Fusion Rule}
The selection of fusion rules largely determines the quality of
fused images \cite{liu2017multifocus}. Existing image fusion algorithms based on deep learning usually calculate the sum of 
the feature maps directly, leaving the information of feature maps not fully mined.

In our method, two complementary fusion rules based on DWT are adopted for wavelet components $C_k$ transformed by feature maps $F_k$, 
including adaptive rule based on regional energy \cite{shen2006region} and \emph{l1-Norm} rule \cite{li2018densefuse},
and the fused wavelet components  are denoted as $F_{r}$ and $F_{l1}$, respectively.
In adaptive rule based on regional energy, different fusion rules are employed for different frequency components, that is, the low-frequency components $C_{kL}$ 
adopts an adaptive weighted averaging algorithm based on regional energy, and for the high-frequency components $C_{kH}$,
 the one with larger variance between $C_{1H}$ and $C_{2H}$ will be selected as the fused high-frequency components. Due to the page limitation,
 the detailed description and futher equations can be found in \cite{fusion}.
 Additionally, to preserve more structural information 
and make our fused image more natural, we apply \emph{l1-Norm} rule \cite{li2018densefuse} to our fusion part, where both low and high 
frequency components are fused by the same rule to obtain  global and general fused wavelet components.

  \section{Experimental  Results and Analysis}

  In this section, to validate the effectiveness and generalization of our WaveFuse, we first compare it with several 
  state-of-the-art methods on four fusion  tasks, including mult-exposure (ME), multi-modal medical (MED), multi-focus (MF) and
  infrared and visible (IV) image fusion. There are 20 pairs of images in each scenario, and all the images are from publicly available datasets \cite{li2018densefuse,VIF,ma2019fusiongan,liu2017multifocus}. For quantitative comparison,
  we use nine metrics to evaluate the fusion results. Then, we evaluate 
  the fusion performance of the proposed method trained with small datasets. Finally, we also conduct
  the fine-tuning experiments on wavelet parameters for the further improvement of fusion performance. 

  \subsection{ Compared Methods and Quantitative Metrics }

  WaveFuse is compared against nine  representative peer methods including discrete wavelet transform (DWT) \cite{li1995DWT}, 
  cross bilateral filter method (CBF) \cite{CBF}, convolutional sparse representation (ConvSR) \cite{liu2016CONVSR}, GAN-based fusion algorithm (FusionGAN) \cite{ma2019fusiongan}, DenseFuse \cite{li2018densefuse}
  IFCNN \cite{11}, and U2Fusion \cite{12}. 
  All the nine comparative methods were implemented based on public available codes, where the 
  parameters were set according to the original papers.

  The commonly used evaluation
  methods can be classified into two categories: subjective evaluation and objective evaluation.
  Subjective  evaluation is susceptible to human factors, such as eyesight, subjective
  preference and individual emotion. Furthermore, no prominent difference among the fusion results
  can be observed in most cases based on subjective evaluation. In contrast, objective evaluation is a relatively accurate and  
  quantitative  method on the basis of mathematical and statistical models.
  Therefore, in order to compare fairly and 
  comprehensively with other fusion methods, we choose the following nine metrics:EN \cite{44}, cross entropy(CE), FMI\_pixel \cite{47}, FMI\_dct \cite{47}, 
  FMI\_w \cite{47}, $\rm{Q^{NICE}}$ \cite{46}), $\rm{Q^{AB/F}}$ \cite{45}, variance(VARI) and subjective similarity (MS-SSIM \cite{ma2015msssim}).
  Each of them reflects different image quality aspects
  , and the larger  the 
  nine quality metrics are, the better the fusion results will be.

  \subsection{ Comparison to Other Methods}
  
  \subsubsection{Subjective Evaluation}
  \begin{table*}[htbp]
    \centering
    \caption{Quantitative comparison of  WaveFuse with existing multi-scene image fusion methods. Red ones are the best results, 
    and blue ones mark the second results.For all metrics, larger is better.}
    \includegraphics[scale=0.6]{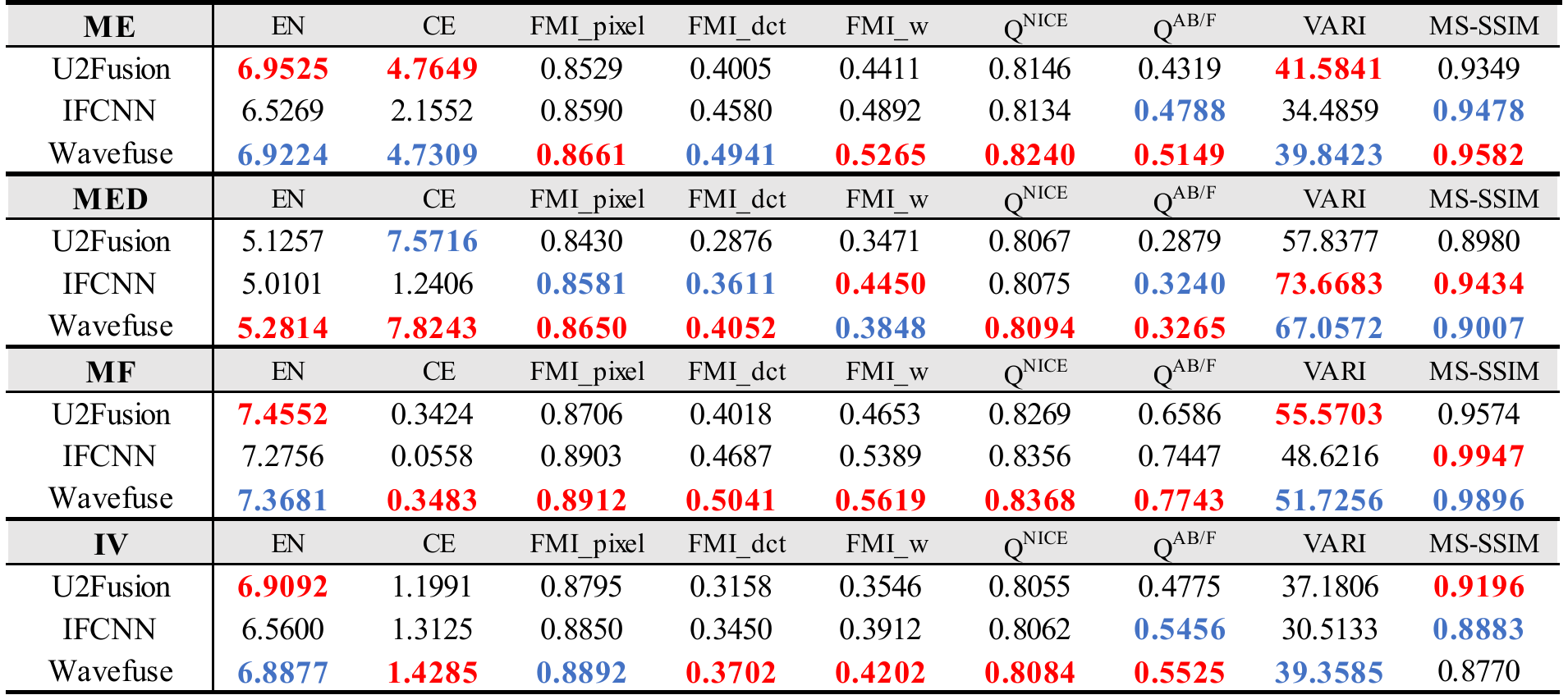}
    \label{tab:multi}
  \end{table*} 
  
\begin{figure*}[htbp]
  \centering
  \includegraphics[scale=0.58]{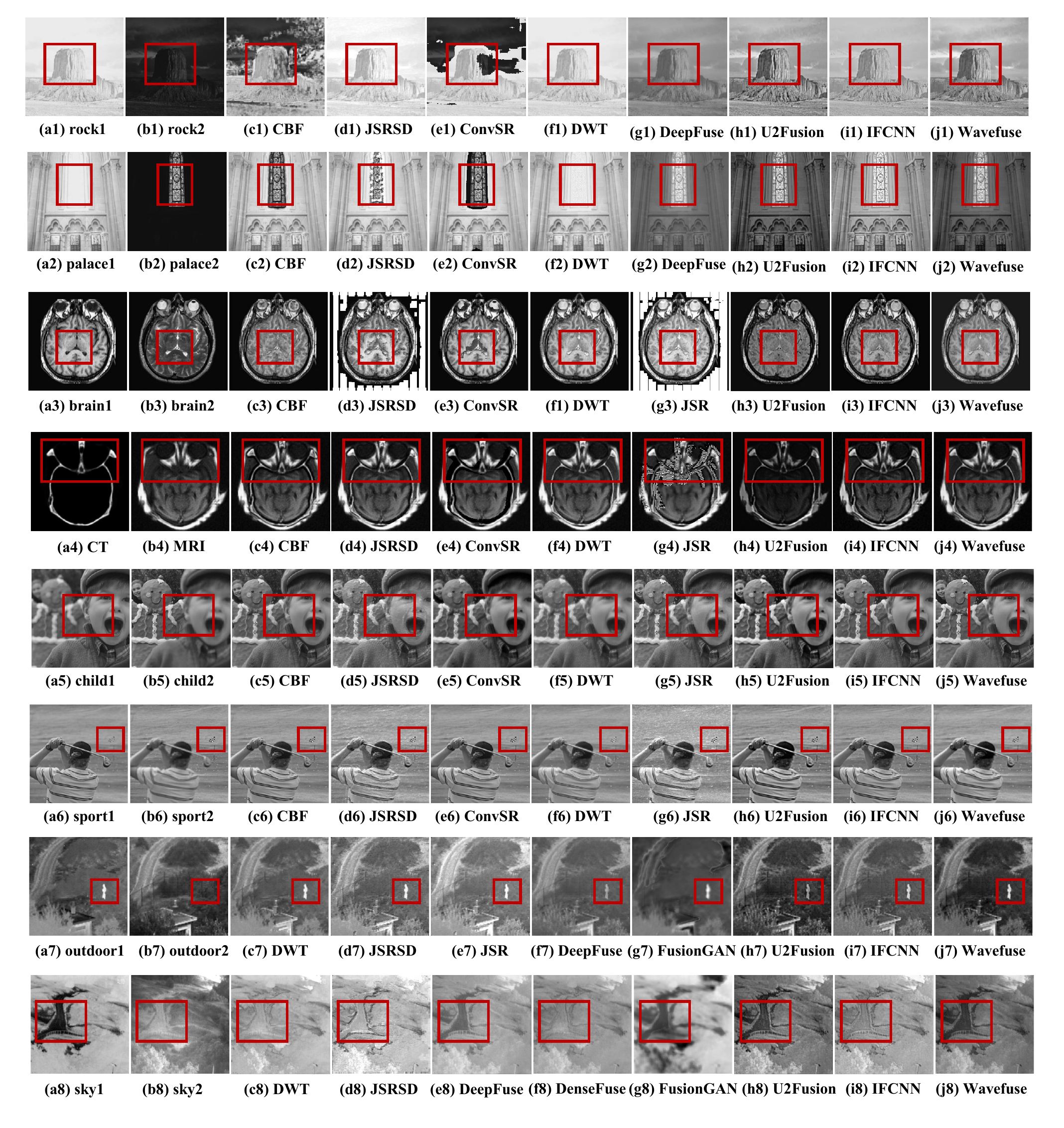}
  \caption{Fusion results by different methods. (a1)-(b1),(a2)-(b2) are two pairs of multi-exposure source images and (c1)-(j1),(c2)-(j2) are the fusion results of them by different methods; 
  (a3)-(b3),(a4)-(b4) are two pairs of multi-modal medical source images and (c3)-(j3),(c4)-(j4) are the fusion results of them by different methods;
  (a5)-(b5),(a6)-(b6) are two pairs of multi-focus source images and (c5)-(j5),(c6)-(j6) are the fusion results of them by different methods;
  (a7)-(b7),(a8)-(b8) are two pairs of infrared and visible source images and (c7)-(j7),(c8)-(j8) are the fusion results of them by different methods;}
  \label{fig:fusion}
\end{figure*}

  Examples of the original image pairs and the fusion results obtained by each comparative method 
for the four scenarios are shown in Fig. \ref{fig:fusion}.

\textbf{Multi-scene Image Fusion:} We first compare the proposed WaveFuse with existing multi-scene image fusion algorithms U2Fusion \cite{12} and IFCNN \cite{11} in 
all the four different fusion scenarios, and the results of the objective metrics are shown in Table \ref{tab:multi}. From Table \ref{tab:multi}, we can see that 
that WaveFuse achieves the best results in almost all scenarios. In some scenarios that it does not achieve the highest metric, our 
method is still close to the highest one. 

\textbf{Multi-exposure Image Fusion:} The multi-exposure image fusion aims to combine different exposures to 
generate better subjective images in both dark and bright regions. From  Fig. \ref{fig:fusion} (c1-j1, c2-j2), we can observe that
CBF and ConvSR generate many artifacts. JSRSD, DWT, DeepFuse and IFCNN suffer from low brightness and blurred details. U2Fusion and WaveFuse achieve 
better fusion reluslts considerding both dark and bright factors.

\textbf{Multi-modal Medical Image Fusion:} Multi-modal medical image fusion can offer more accurate and 
effective information for biomedical research and clinical applications. Better multi-modal medical fused image 
should provide combined features sufficiently and preserve both significant textural features. As shown in 
Fig. \ref{fig:fusion} (c3-j3, c4-j4), JSR, JSRSD and ConvSR shows obvious artifacts in the whole image. DWT and CBF fail to 
preserve the crucial features of the source images. U2Fusion shows better visual results than  the above-mentioned
methods. However, DenseFuse still weakens the details and brightness. Information-rich fused images can be obtained by IFCNN.
In contrast, our method preserves  the details and edge information of both source images, which is 
more in line with the perception characteristics of the human vision compared to  other fusion methods.

\textbf{Multi-focus Image Fusion:} The multi-focus image fusion aims to reconstruct a fully focused image from 
partly focused images of the same scene. From  Fig. \ref{fig:fusion} (c5-j5, c6-j6), we can observe that JSR and JSRSD shows obvious blurred artifacts. 
DWT shows low brightness in the fusion results. Other compared methods perform well. 

\textbf{Infrared/Visible Image Fusion:} Visible images can capture more detail information compared
to  infrared images. However, the interested objects can not be easily observed in visible image especially
when  it is under low contrast circumstance and the light is insufficient. Infrared images can provide 
thermal radiation information, making it easy to detect the salient object even in  complex background. Thus, the 
fused  image can provide more complementary information. Fig. \ref{fig:fusion} (c7-j7, c8-j8) show fusion results of  infrared and visible
images with the comparison methods. JSR, JSRSD  and FusionGAN exhibit significant artifacts., and U2Fusion shows unclear salient 
objects. The results in  DWT, DenseFuse and IFCNN weaken the contrast. 
We can see that,  WaveFuse preserves more details in high contrast and brightness.
\begin{table*}[t]
  \centering
  \caption{The average values of fusion quality metrics for fused images of four different scenarios. Red ones are the best results, 
  and blue ones mark the second results. For all metrics, larger is better. }
  \includegraphics[scale=0.6]{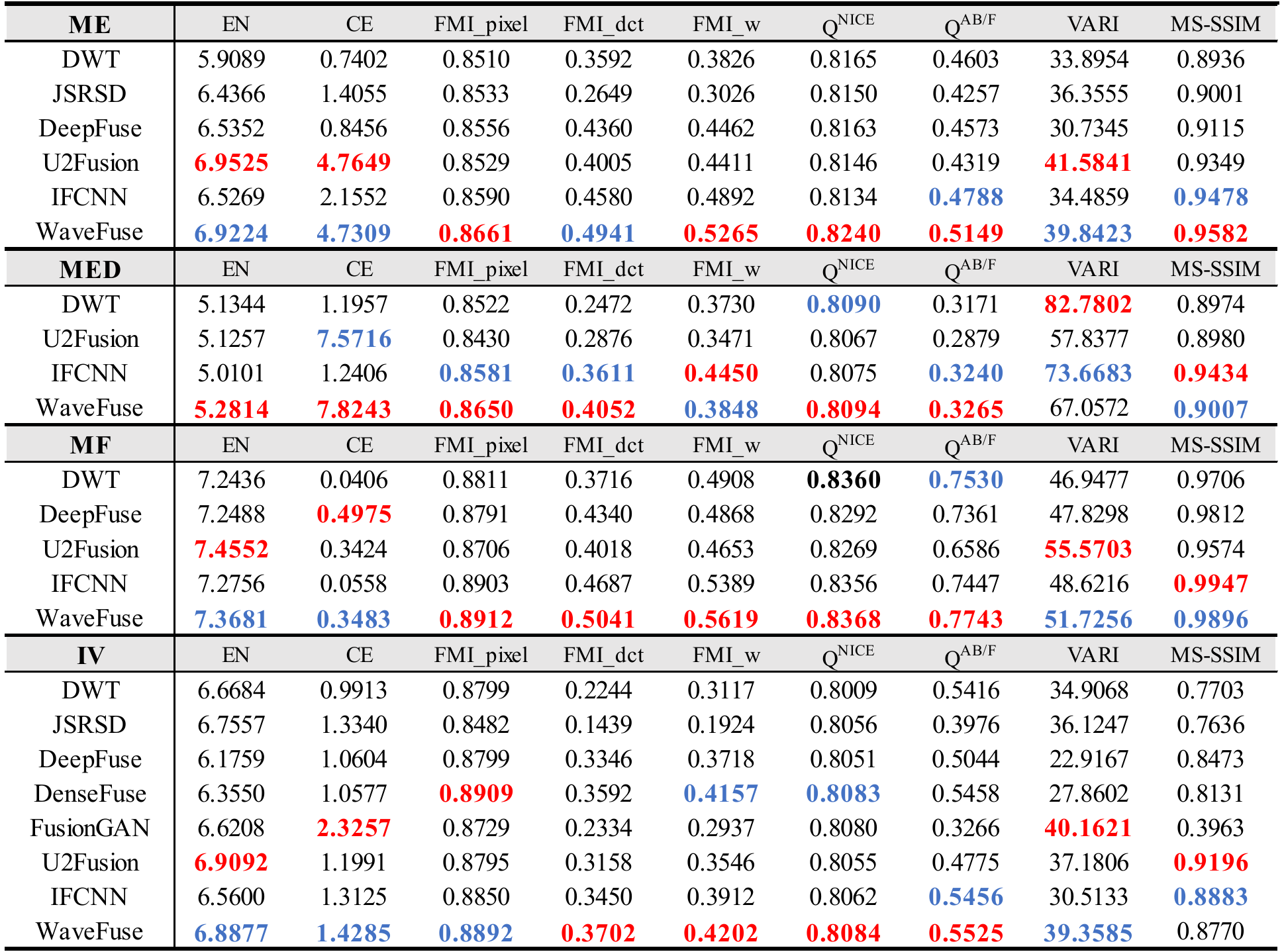}

  \label{fig:fusion_all}
\end{table*}
  \subsubsection{Objective Evaluation}

  From Fig. \ref{fig:fusion}, we can observe
  that the fusion results of CBF and ConvSR in multi-exposure images, the fusion results of CBF, JSR, JSRSD and ConvSR in 
   multi-modal medical images and  the fusion results of JSRSD in multi-focus images contain poor visual 
  effects owing to considerable artificial noise, and in this case their objective quality metrics will not be calculated for the 
  quantitative evaluation.

  Table \ref{fig:fusion_all} shows the average values of the fusion quality metrics among 
  four different fusion tasks by different fusion methods. In multi-exposure image fusion, 
  our method ranks first in  FMI\_pixel, FMI\_dct, FMI\_w, $\rm{Q^{NICE}}$, $\rm{Q^{AB/F}}$  and MS-SSIM, 
  and ranks second in EN, CE and VARI. In multi-modl medical  image fusion, 
  our method ranks first in EN, CE, FMI\_pixel, FMI\_dct, $\rm{Q^{NICE}}$ and  $\rm{Q^{AB/F}}$, 
  and ranks second in  FMI\_w and  MS-SSIM.  In multi-focus image fusion, 
  our method ranks first in FMI\_pixel, FMI\_dct, FMI\_w, $\rm{Q^{NICE}}$ and $\rm{Q^{AB/F}}$, and ranks second in EN, CE, VARI and  MS-SSIM . 
  In the infrared and visible image fusion,
  our method obtains the highest metrics in FMI\_dct, FMI\_w, $\rm{Q^{NICE}}$ and $\rm{Q^{AB/F}}$, and  ranks second in EN, FMI\_pixel and VARI.   Furthermore, from the value of the last row among three image fusion task in 
  Overall, compared with other peer methods, our proposed method  achieves the highest values in most fusion quality metrics and the second in the remaining metrics.




  \subsection{ Comparison of Using Different Training Dataset}

  In order to further demonstrate the effectiveness and robustness of our network, 
  we conducted  experiments on another three different training minisets: MINI1-MINI3, each of which 
  contains 0.5\%, 1\% and 2\% images respectively chosen randomly from COCO, and the fusion results 
  difference can be found among the subjective fused images, we compared the objective fusion results of 
  WaveFuse on COCO and MINI1-MINI3. The fusion performance was compared and analyzed by the averaged 
  fusion quality metrics. In WaveFuse, higher performance is even achieved by training on minisets. Due to page limitation, further details can be found in \cite{fusion}.

   Furthermore,  we can observe that
  WaveFuse is trained on minisets within one hour, where the GPU memory utilization is just 4085 MB, so 
  it can be trained with lower computational cost compared with that trained in COCO (7.78h and 17345 MB). Accordingly, we can learn that our proposed 
   network is robust both to the size of the training dataset and to the selection of training images.

  

  \subsection{Ablation  Studies}

  $\bullet$ \textbf{DWT-based feature fusion}
  In this section, we attempt to explain why DWT-based feature fusion module can improve fusion performance.
  DWT has been a poweful multi-sacle analysis tool in signal and image processing  since it was proposed. 
  DWT transforms the images into different low and high frequencies, where low frequencies represent contour and edge information and high
  frequencies represent detailed texture information\cite{li1995DWT}. In this way, DWT-based fusion methods first transform the images into low and high 
  frequencies, and then fuse them in the wavelet domain, achieving promising fusion results. Inspired by DWT methods, we apply DWT-based fusion module
  to deep feature fusion extracted by deep-learning models, so as to fully utilize the information contained in deep features. We conducted the ablation 
  study about DWT-based  feature fusion module, and the results is shown in Table \ref{tab:dwt}. As we can see, when we apply the module, the fusion performance is 
  indeed improved largely.
  
  \begin{table}[hbtp]
    \centering
    \caption{Ablation study on the DWT-based feature fusion module. Red ones are the best results. For all metrics, larger is better.}
    \includegraphics[scale=0.75]{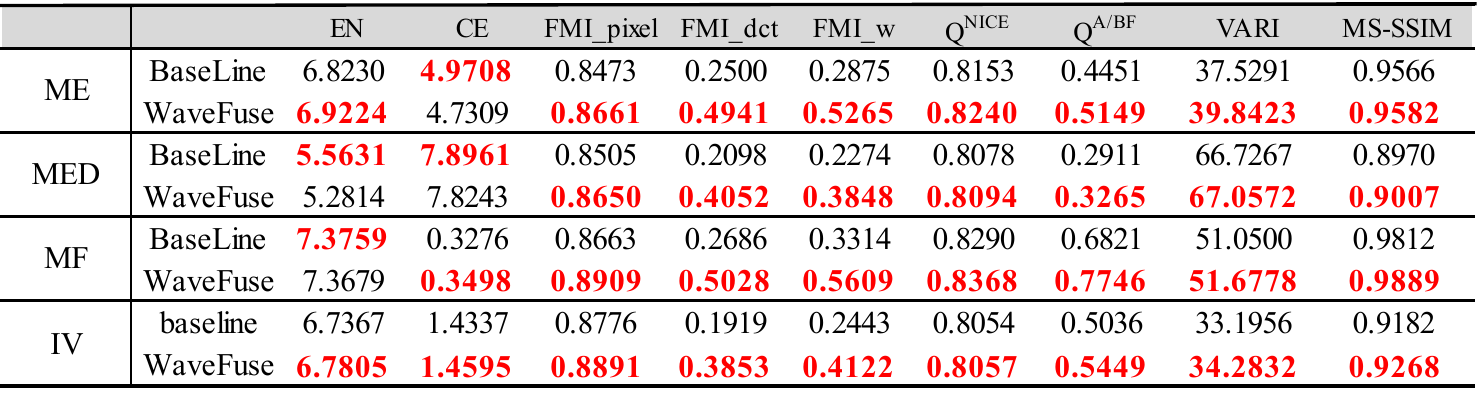}
     \label{tab:dwt}
  \end{table} 

  $\bullet$ \textbf{Experiments on Different Wavelet Decomposition Layers and Different Wavelet Bases}
  In wavelet transform, the number of decomposition layers and the selection of different wavelet 
  bases could exert great impacts on the effectiveness of wavelet transform. We also conducted the ablation study on different settings.
  Due to the page limitation, we just give our final conclusion, when  the number of decomposition layers and the  wavelet 
  base are set as 2 and \emph{db1} respectively, we achieve the best fusion results. More details can be found in \cite{fusion}.

  \section{Conclusions}
  In this paper, we propose a novel image fusion method  through the combination of a multi-scale discrete wavelet 
  transform based on regional energy and deep learning. To our best knowledge, this is the first time 
  that a conventional technique is integrated for feature maps fusion in the pipeline of deep learning based image fusion 
  methods, and we think there are still a lot of possibilities to explore in this direction. 
  
  Our network consists of three parts: an encoder, a DWT-based fusion part and a decoder. The 
  features of the input image are extracted by the encoder, then we use the adaptive fusion rule at 
  the fusion layer to obtain the fused features, and finally reconstruct the fused image through the 
  decoder. Compared with existing fusion algorithms, our proposed method achieves better 
  performance. Additionally, our network has strong universality and can be applied to various  image fusion
  scenarios. At the same time, our network can be trained in smaller datasets to obtain the comparable 
  fusion results trained in large datasets with shorter training time and higher efficiency, alleviating the dependence
  on large datasets. Extensive experiments on different
  wavelet decomposition layers and bases demonstrate the possibility of further improvement of our method. 
%
%
%
\bibliographystyle{splncs04}
\bibliography{mybibfile}

\end{document}